\pdfoutput=1

\documentclass[11pt]{article}

\usepackage{emnlp2021}

\usepackage{times}
\usepackage{latexsym}
\usepackage{graphicx}
\usepackage{multirow}

\usepackage[T1]{fontenc}

\usepackage[utf8]{inputenc}

\usepackage{microtype}
\usepackage{graphicx}
\usepackage[shortlabels]{enumitem}
\usepackage{amsmath}
\usepackage{amssymb}

\DeclareMathOperator{\edits}{edits}

\title{Minimal Supervision for Morphological Inflection}

\author{Omer Goldman \\
  Bar Ilan University \\
  \texttt{omer.goldman@gmail.com} \\\And
  Reut Tsarfaty \\
  Bar Ilan University \\
  \texttt{reut.tsarfaty@biu.ac.il} \\}

\begin{document}
\maketitle
\begin{abstract}

Neural models for the various flavours of morphological inflection tasks have proven to be extremely accurate given ample labeled data, yet labeled data may be slow and costly to obtain. In this work we aim to overcome this annotation bottleneck by bootstrapping labeled data from a seed as small as {\em five} labeled inflection tables, accompanied by a large bulk of unlabeled text.
Our bootstrapping method exploits the orthographic and semantic regularities in morphological systems in a two-phased setup, where word tagging based on {\em analogies} is followed by word pairing based on {\em distances}.
Our experiments with the Paradigm Cell Filling Problem over eight typologically different languages show that in languages with relatively simple morphology, orthographic regularities on their own allow inflection models to achieve respectable accuracy. Combined orthographic and semantic regularities alleviate difficulties with particularly complex morpho-phonological systems. We further show that our bootstrapping methods substantially outperform hallucination-based methods commonly used for overcoming the annotation bottleneck in morphological inflection tasks.

\end{abstract}

\section{Introduction} \label{sec:intro}
The introduction of neural models into natural language processing in the last decade has led to huge improvements in all {\em supervised} generation tasks, including morphological %
inflection.\footnote{In recent years the term \textit{reinflection} has surfaced as a reference to morphological inflection done not necessarily from the lemma. In this paper we will refer to both inflection and reinflection as "inflection", and specify whenever we refer to inflection done exclusively from the lemma.} %
In particular, previous works \cite{cotterell-etal-2017-conll, silfverberg-hulden-2018-encoder} have achieved near-perfect performance over the Paradigm Cell Filling Problem (PCFP)  \cite{ackerman-etal-2009-parts}, wherein models are required to provide any form in an inflection table, given a few forms of the same lexeme.\footnote{Throughout the paper we  conform to the linguistic terminology, where `{lexeme}' stands for an abstract lexical entry, e.g., the English \textsc{run}, and its `forms' are the words that convey this lexical meaning with some inflectional relations between them, e.g. \textit{run}, \textit{running} but not \textit{runner}. `{Paradigm}' will stand for a group of lexemes sharing a POS tag, in the same manner as the English lexeme \textsc{run} is part of the verbal paradigm.} %

Two lines of recent work made progress towards less supervision, in different fashions. The first simply %
provided scenarios with smaller training sets --- for example, in SIGMORPHON's shared tasks \cite{cotterell-etal-2017-conll, cotterell-etal-2018-conll}. The second research avenue %
aims to discover the paradigmatic structure of an unknown language given a large bulk of unlabeled data, either alone \cite{soricut-och-2015-unsupervised, elsner-etal-2019-modeling}, accompanied by a list of all relevant forms in the vocabulary \cite{erdmann-etal-2020-paradigm}, or by a list of lemmas \cite{jin-etal-2020-unsupervised}.

The problem with the first kind of attempts is that given the neural nature of the most successful models, their performance on limited supervision is capped, and data augmentation is likely to help only if the initial data is diverse enough. As for the second scenario of no supervision at all, it is somewhat pessimistic and unrealistic. Even if much labeled data for a language does not exist for a low-resourced language, typically there exists {\em knowledge} about its paradigm structure that can be employed. %
UniMorph \cite{kirov-etal-2018-unimorph}, for example, includes small amounts of labeled inflection tables for many languages, from obscure ones like Ingrian to national languages with widespread usage that lack global attention like Georgian.

In this work we propose a new, {\em low-resourced 
morphological inflection} scenario, which is more optimistic and realistic for those widely-spoken sparsely-annotated languages. %
We assume a {\em minimal} supervision set and a large bulk of unlabeled text, thus balancing both trends of lowering supervision resources. %
 We bootstrap a tiny amount of as little as {\em five} inflection tables, that could be easily written by hand, into a full-blown training set, albeit noisy, for an inflection model trained in a supervised manner.
 Our approach makes use of the regularities abundant in inflectional morphology, both  orthographic regularities and semantic ones.

Based on this method we train morphological inflection models for eight languages and show that, for the five Indo-European of them,  orthographic regularity is enough to train a morphological inflector that achieves reasonable success.
We further show that for languages with complicated morpho-phonological systems, such as Finnish, Turkish and Hungarian, a method combining both orthographic  and semantic regularities is needed in order to reach the same level of performance. %
An error analysis reveals that the closer an inflection is to the verge of disappearance, the poorer our system performs on it, as less examples exist in the data-derived vocabulary.
Our models outperform \citet{makarov-clematide-2018-neural}'s model designed for low-resourced setting, even when equipped with additional hallucinated data \cite{anastasopoulos-neubig-2019-pushing}. We also outperform the best model of \newcite{jin-etal-2020-unsupervised},  that didn't use any inflection tables, and their skyline for most languages.

We conclude that bootstrapping datasets for inflectional morphology in low-resourced languages, is a viable strategy to be adopted and   explored.\footnote{Our code is available on \url{https://github.com/OnlpLab/morphodetection}.}

\section{The Minimally-Supervised Setup}
\label{sec:setup}

\paragraph{Problem Statement}
Let \(\mathcal{L}\) be a set of lexemes, each corresponding to an inflection table $W^l=\{w_{t_1}^l...w_{t_m}^l\}$ where $w_{t_i}^l$ is a form of lexeme $l$ bearing the feature bundle $t_i$.
Our goal is to train an inflection model that maps words bearing one set of features to words bearing another set within the same lexeme.
\[ (  \langle t_j, w_{t_j}^{l_i}\rangle , t_k  )  \mapsto w_{t_k}^{l_i} 
\]
For example, in the French verbal paradigm:
\begin{tabular}{r@{}c@{}l}
     \big(\(\langle\) \textsc{IndPrs1Sg}, \emph{finis} \(\rangle \),  \textsc{SbjvPst2Sg}&\big)&\\
    $\mapsto$ \emph{finis}&\emph{s}&\emph{es}
\end{tabular}

In order to  induce this function, we propose a {\em minimally-supervised} scenario where we are only given a small set of 
$n$ examples of complete inflectional tables  $\mathcal{L} = \{W^{l_1}...W^{l_n}\}$  (each of which is of size \(m\)), and a large bulk of naturally occurring (that is, in-context) unlabeled data in that language, that is, \(w_1 ..... w_c\), such that \(c >> nm\). 

\section{The Algorithmic Framework}
\label{sec:framework}

This work suggests utilizing the patterns exhibited  in the small supervision seed, and finding words  that exhibit  {\em similar} (or, {\em analogous}) patterns. 

The algorithm proposed here works in two phases. First we tag words  with morphological features if they are found similar (analogous) to examples in the minimal supervision.
Then, we pair the tagged words such that each pair will include two forms of the same lexeme. This algorithmic division of labor allows the pairing module to provide a sanity check, and reduce the noise potentially lingering from the word-tagging module.

The above sketch of the algorithm is quite generic, that is, we can get different instantiations of this framework by plugging in different ways to calculate {\em similarities} or {\em analogies} between words.
In our various algorithmic implementations, we will use the  regularities that prevail in inflectional morphology and are detectable even from a tiny amount of supervision. These regularities are manifested both orthographically, as edits between forms tend to repeat themselves across lexemes, and semantically, as forms that share an inflectional function tend to be used similarly. %

In the rest of this section we will describe both modules, each with its different variants depending on the different notion of {\em similarity} used.

\subsection{Morphological Features Assignment}
\label{algo1}

In order to assess whether a pair of unseen words $w_1, w_2$ belong to the same lexeme, we first need to characterize the relationship between those two words. 
It is then imperative to compare the concrete relation between \(w_1,w_2\) to some representation of the abstract relation $R_{t_j,t_k}$ between $2$ morphological categories $t_j$ and $t_k$ in the same paradigm. If these concrete and abstract relations are sufficiently similar, $w_1$ and $w_2$ will be tagged as bearing features $t_j$ and $t_k$, respectively.
The idea, in a nutshell, is to obtain the representation $R_{t_j,t_k}$  by aggregating differences between the forms of the \(t_j,t_k\) entries in all  $n$  inflection tables in the minimal seed. These differences  can be stated in terms of either semantics, orthography or a combination thereof. %

\subsubsection{Orthography-Based Tagging}
\label{algo_orth}

In the orthographic case we define the difference between a pair of words as the edits needed to get from one word to the other. The edits our system expects are a list of sub-strings that were deleted and a list of those added, sorted from left to right.\footnote{Though without position indices, in order to deal with words of various lengths.}

For every pair of morphological categories $(t_j, t_k)$, their orthographic relation $R_{t_j,t_k}^{O}$ is defined by the set of edits observed in the $n$ supervision inflection tables, one from each lemma $l_i$
\[ \hat{R}_{t_j,t_k}^{O} := \big\{\edits (w_{t_j}^{l_i}, w_{t_k}^{l_i}) \big\}_{i=1}^n \]
\noindent In order to check whether a pair of new words $(w_a, w_b)$ exhibits a  relation between two   \(t_j,t_k\) morphological categories, we check whether the edits between them belong to  the relation representation:
\[ \edits (w_a, w_b) \stackrel{?}{\in} \hat{R}_{t_j,t_k}^{O} \]

Consider the example from the French verbal paradigm where the examples for the relation (\textsc{IndPrs1Sg}, \textsc{IndPrs3Sg}) are \{\textit{(finis,finit), (bois, but), (parle,parla)}\}. In this case the representation of this relation is:
\begin{align*}
\hat{R}_{\textsc{IndPrs1Sg},\textsc{IndPrs3Sg}}^{O}& =\\ 
\{(\textit{`s',`t'}), (\textit{`ois'}&\textit{,`ut'}), (\textit{`e',`a'})\}
\end{align*}

Since the edits between (\textit{aime, aima}) are identical to those between (\textit{parle,parla}), then \textit{aime} would be correctly considered for tagging as \textsc{IndPrs1Sg} and \textit{aima} -- as \textsc{IndPstPrf3Sg}.

This procedure is however highly prone to coincidence.
For example, the edits between \textit{le} and \textit{la} are  the same as between \textit{parle} and \textit{parla}, although the former are actually determiners, and not part of the verbal paradigm. Given the multitude of relations available, we can expect many edits between {\em incidental} pairs of words to match an edit seen in the gold data. 
To overcome this, we propose to take %
the complete paradigm structure into account, rather than considering word pairs in isolation.

Concretely, we propose to tag only words that have been found to answer this criterion for {\em multiple relations in the same paradigm}, covering at least half of the size of the paradigm. So \textit{aime} would be tagged as \textsc{IndPrs1Sg} since $\edits(\textit{aime, aima}) = \edits(\textit{parle,parla})$ and $\edits(\textit{aime, aimons}) = \edits(\textit{parle, parlons})$ and so on, but \textit{le} won't be tagged as  \textsc{IndPrs1Sg} since the French vocabulary does not contain \textit{lons}. %

With this tightened criterion,  the orthographic algorithm might be sufficiently precise, but may not be sufficiently diverse, so as to obtain high recall.

\subsubsection{Semantics-Based Tagging}
\label{algo_sem}
The orthographic criterion only considers exact match with the observed edits so it is expected to miss %
irregulars and classes unattested in the $n$ supervision inflection tables.\footnote{the term `class' refers to a group of lexemes in a paradigm that display similar inflection patterns. Traditionally known as `conjugation' and `declension' in the description of verbal and nominal paradigms, respectively. E.g., the Spanish verbal paradigm is said to include $3$ classes: \textit{-er}, \textit{-ar} and \textit{-ir} verbs.} This can pose a significant problem to paradigms that have more than $n$ classes or display significant morpho-phonological processes %
not present in the labeled examples.

To overcome the generalization problem of orthographic edits we propose to consider semantic regularities,
since the differences in meaning and usage rarely correlate with orthography. Semantic regularity arises from agreement, a phenomenon in which words in a sentence must have the same morphological features as some other words in the sentence, effectively creating equivalence classes.
Modern algorithms for word embeddings that extract semantics from co-occurrences, following \newcite{firth-1957-synopsis}, are naturally suitable to exploit this kind of regularity.

In this setting the difference between words is defined as the difference between their embedded vectors. And for every pair of morphological feature bundles $(t_j, t_k)$ the representation of their semantic relation is estimated by the average over those relevant examples
\[ \hat{R}_{t_j,t_k}^{S} = \frac{1}{n} \sum_{i=1}^n v(w_{t_j}^{l_i}) - v(w_{t_k}^{l_i}) \]

A new word pair will be tagged \(t_j,t_k\) if their difference is close enough, in cosine-distance terms, to $\hat{R}_{t_j,t_k}^{S}$
\[ D_C \big(\hat{R}_{t_j,t_k}^{S}, v(w_a)-v(w_b) \big) \stackrel{?}{\leq} \hat{C}_{t_j,t_k}^{S} \]
\noindent where $D_C$ is the cosine distance function and $\hat{C}_{t_j,t_k}^{S}$ is an estimation of a relation-specific cut-off score set by the average scatter of the relevant supervision examples around their average:
\[ \hat{C}_{t_j,t_k}^{S} = \frac{1}{n} \sum_{i=1}^n D_C \big( \hat{R}_{t_j,t_k}^{s}, v(w_{t_j}^{l_i}) - v(w_{t_k}^{l_i}) \big) \]

Although lacking the orthographic disadvantages, here $\hat{R}_{t_j,t_k}^{S}$ might be a biased representation that misses many examples, or mistakenly tags  incorrect words. For this reason we suggest the third algorithm combining both types or regularities.

\subsubsection{Combined Tagging}
\label{algo_comb}

The idea behind this variant is to consider word pairs that answer both the orthographic and semantic criteria as \textit{semi-gold} examples that will be added to better estimate the relation $\hat{R}_{t_j,t_k}^{S}$. Thus we harness the accuracy of the orthographic criterion to combat the bias in the semantic representation.

Specifically, a pair of words $(w_1, w_2)$ is considered \textit{semi-gold} if their edit script is in $\hat{R}_{t_j,t_k}^{O}$, \textbf{and} the distance of their difference vector from the relation vector $\hat{R}_{t_j,t_k}^{S}$ is smaller than the furthest corresponding difference vector of gold examples.

Note that we relaxed both criteria comparing to those in the previous sections. The orthographic criterion is relaxed by dropping the requirement for the complete paradigm, and the semantic criterion is relaxed by replacing $\hat{C}_{t_j,t_k}^{S}$ with a more inclusive cut-off relying on max rather than average distance%
\[ \hat{C}_{t_j,t_k}^{Comb} = \max_i D_C \big( \hat{R}_{t_j,t_k}^{s}, v(w_{t_j}^{l_i}) - v(w_{t_k}^{l_i}) \big) \]

The relaxations allow inclusion of more \textit{semi-gold} examples, and they are sensible as both criteria are mutually constraining. %

Pairs of words that have satisfied both criteria are very likely correct examples of the morphological relation $(t_j,t_k)$. Thus they are added to create a new semantic representation $\hat{R}_{t_j,t_k}^{S}$. The new representation may be used again to find more semi-gold examples, executing this stage iteratively. We set the stopping criterion when either no examples are added to $R_{t_j,t_k}^{S}$ after an iteration, or when so many examples where added so that the run time per iteration exceeded $48$ hours in our implementation. 

Once $\hat{R}_{t_j,t_k}^{S}$ is settled, and in order to include tagging of words with different edits, we finally tag words according to the semantic criterion alone with the corresponding $\hat{C}_{t_j,t_k}^{S}$.

\subsection{Pairing Tagged Words}

Given the output of the first module, a list of words tagged with some morphological sets of features, the second module pairs tagged words from the first module,  such that both tagged words are forms of the same lexeme. 

We start by grouping the tagged words from the first module into \(m\) bins $\{B_{t_i}\}_{i=1}^m$, each containing all words tagged with $t_i$. We then need to find for each word $w_{t_i}^{l_j} \in B_{t_i}$ corresponding words from other bins that are forms of the same lexeme $l_j$.

Ideally, if each bin contained exactly one correct form of every lexeme, pairing the most similar words across bins should suffice in order to collect all the forms in the paradigm.
In reality, due to noise, bins may include several forms of the same lexeme or none at all. Assuming enough words are tagged, it seems better to simply drop the cases where several forms of the same lexeme are occupying the same bin, rather then trying to locate  the one that was tagged correctly. Therefore, we would like to pair words such that they are \textit{distinct} nearest neighbors across bins, i.e., where  the second nearest neighbor is much more distant.

To achieve this, we scale the distance using the Cross-domain Similarity Local Scaling (CSLS) score, suggested by \newcite{conneau-etal-2018-word} in the context of bi-lingual embeddings.
The CSLS score scales the cosine distance between vectors across groups $x\in X, y \in Y$ with the average distance from their top-$k$ nearest neighbors.
We set $k=2$ when applying this method here, since we aim  to discard cases where a form $w_{t_i}^{l_q} \in B_{t_i}$ has {\em at least} two corresponding forms in another bin $w_{t_j}^{l_q},  w_{t_k}^{l_q}\in B_{t_j}$, with one presumably misplaced. The definition of CSLS for \(k=2\) can be written as 
  \[
    \begin{aligned} 
    CSLS & (x,y) = D(x,y) - \\
    &\frac{1}{2} D(x,NN_2^x(Y)) - \frac{1}{2} D(NN_2^y(X), y)
    \end{aligned}
  \] 
\noindent where $NN_2^x(Y)$ is $x$'s second nearest neighbor in group $Y$. 

Although the original distance measure $D(\cdot , \cdot)$ used by \newcite{conneau-etal-2018-word} is the cosine distance function, any distance measure will do.
In the semantic algorithm (Sec. \ref{algo_sem}), we apply cosine distance that reflects semantic similarity. In the orthgraphic algorithm (Sec. \ref{algo_orth}), we plug in the Levenshtein edit distance, to utilize the orthography rather than the semantics.

After scoring all possible word pairs, the best scored pairs are taken as a training set for a supervised neural morphological inflector.

\section{Experimental Evaluation} \label{sec:experiments}

\subsection{Experimental Setup}
\label{sec:exp_setup}

We set out to empirically examine whether it is possible to train a morphological inflection model while starting with a minimal number of labeled inflection tables, using regularities of different types. Our experiments include the verbal paradigm of eight languages: English, Russian, Latvian, Spanish, French, Finnish, Hungarian and Turkish, although our methods are suitable for any paradigm. %

To make evaluation of the inflection model possible, we had to choose languages with a sufficient amount of gold (test)  data, and \textit{simulate} a low-resource scenario for training. This limited us to western languages, and we aimed to include as many non-Indo-European languages as possible.

\paragraph{The Unlabeled Data} The problem setting we specified in Section~\ref{sec:setup} designates the use of a bulk of unlabeled text. In actuality, the proposed algorithm makes use of the text  for (i) collecting a vocabulary to seek tagging candidates, and (ii) training embeddings to be used for calculating the semantic relations between candidate pairs.
In our experiments, we simply employed language-specific pre-trained word embeddings  for both purposes. 
 
We used the pre-trained FastText vectors provided by \newcite{grave-etal-2018-learning}, following their reported success over morphological analogies in the Bigger Analogy Test Set \cite{li-etal-2018-subword}. We clipped the $2$ million long FastText vocabulary to include only real words, i.e., we keep only lower-cased tokens that do not include non-alphabetic characters, and include at least one vowel.\footnote{This last criterion happens to be suitable for all 8 languages examined. Admittedly there are languages, like Czech, that allow less sonorant syllable nucleus.} 
This procedure downsized the vocabulary size to between about $200k-500k$ words per language.\footnote{To reduce run time the Finnish and Hungarian vocabularies were further reduced to include only words that appeared at least twice in the Finnish Parole and Hungarian Gigaword corpora \cite{bartis-1998-parole, oravecz-etal-2014-hungarian}.} Additionally, for run time reasons, we capped the size of the vocabulary for the orthographic variant to $200k$ words per language.

\paragraph{Minimal Supervision Source} For every language we extracted inflection tables for $5$ lexemes from UniMorph \cite{kirov-etal-2018-unimorph}. We aimed for lexemes that are both frequently used, to have robust embedded representations, and from diverse classes, to capture as much as possible of the linguistic behaviors of the language. To this end we targeted lexemes with the largest amount of forms appearing in the embeddings' vocabulary and manually selected from them such that they belong to as diverse classes as possible.\footnote{Cases of syncretism were trivially solved by merging $2$ categories into one category with $2$ tags if all supervision lexemes had the same forms in those $2$ categories. This strategy might pose a problem in cases of \textit{partial} syncretism, where categories differ in forms only in classes that happen to be absent from the minimal supervision. The manually enforced class diversity in the selection process was devised to solve this problem as well.}

\paragraph{The Inflection Model} As our supervised inflection model, to be trained on the bootstrapped data, we used an out-of-the-box sequence-to-sequence LSTM model with attention over the input's characters and the  features of the source and target forms. We used the publicly available model of \newcite{silfverberg-hulden-2018-encoder}. We trained the model for 50 epochs, on either up to $10,000$ data training examples outputted by our system in the minimally supervised scenario, or on $6,000$ training examples in the supervised scenario. The latter provides an upper-bound for the performance of our minimally supervised system. \newcite{silfverberg-hulden-2018-encoder} provided train and test data for $4$ of the languages examined here, and we hand-crafted  similar data sets for English, Russian, Hungarian and Turkish.

The model's evaluation metric is exact match of the outputted string to the gold output.

\paragraph{Models, Baseline and Skyline}
We report results for the three system variants we tested: %
\begin{itemize}[nosep,leftmargin=0.4cm]
    \item \textsc{Orth}: The orthography-based system (\S\ref{algo_orth});
    \item \textsc{Sem}: The embedings-based system (Sec.\ \ref{algo_sem});
    \item \textsc{Comb}: The combined system (Sec.\ \ref{algo_comb}).
\end{itemize}
\noindent In addition, in order to assess the added value in our systems we include accuracy for two baseline models trained in the low-resourced setting without finding new exmaples:
\begin{itemize}[nosep,leftmargin=0.4cm]
    \item \textsc{Overfit}: Our sequence-to-sequence inflection model, based on \newcite{silfverberg-hulden-2018-encoder}.
    \item \textsc{CA}: The neural transition-based model model of \newcite{makarov-clematide-2018-neural}.\footnote{This model was designed for low-resourced settings and achieved state of the art results in SIGMORPHON's $2018$ shared-task on inflection from lemma \cite{cotterell-etal-2018-conll}. We adapted it to allow for inflection from an arbitrary form.}
\end{itemize}

We also include the performance of a model trained in a fully-supervised fashion (\textsc{Sup}) as an upper-bound.

\begin{table*}[t]
\small{
 \begin{center}
 \begin{tabular}{l|c|c|c|c|c|c|c|c|c}
 \textbf{Inflection} & Average & \textsc{Eng} & \textsc{Rus} & \textsc{Lav} & \textsc{Fra} & \textsc{Fin} & \textsc{Spa} & \textsc{Tur} & \textsc{Hun} \\ %
 \hline
 \hline
 \textsc{Overfit} & 0.01 & 0.00 & 0.00 & 0.01 & 0.01 & 0.00 & 0.00 & 0.01 & 0.00 \\ 
 \textsc{CA} & 0.18 & 0.38 & 0.30 & 0.15 & 0.06 & 0.06 & 0.02 & 0.22 & 0.23 \\
 \hline
 \textsc{Orth} & 0.40 & 0.86 & \textbf{0.75} & \textbf{0.49} & 0.31 & 0.13 & 0.50 & 0.32 & 0.01 \\ %
\textsc{Sem} & 0.09 & 0.01 & 0.33 & 0.01 & 0.07 & 0.00 & 0.24 & 0.03 & 0.00 \\ %
 \textsc{Comb} & \textbf{0.54} & \textbf{0.90} & 0.73 & 0.48 & \textbf{0.32} & \textbf{0.48} & \textbf{0.53} & \textbf{0.49} & \textbf{0.48} \\ %
 \hline
 \hline
 \textsc{Supervised} & 0.93 & 0.95 & 0.93 & 0.86 & 0.84 & 0.95 & 0.95 & 0.94 & 0.98 \\ %
  \hline
 \multicolumn{2}{c|}{Paradigm size} & 5 (8) & 15 (15) & 27 (34) & 32 (48) & 39 (39) & 56 (65) & 30 (30) & 48 (49) %
 \end{tabular}
 \end{center}
 }
\caption{Morphological inflection accuracies for all languages and systems. Paradigm sizes in number of forms (functions) is also included for reference. The best minimally-supervised results are in \textbf{bold}.}
\label{tab:main_res}
\end{table*}

\begin{table}[t]
\small{
\begin{center}
\begin{tabular}{l|c|c|c|c|c|c}
 & Avg & \textsc{Eng} & \textsc{Rus} & \textsc{Fin} & \textsc{Spa} & \textsc{Tur} \\
 \hline
 \hline
 \textsc{CoNLL17} & 0.49 & 0.71 & 0.37 & 0.25 & \textbf{0.76} & 0.35 \\

 \textsc{PCS} 
 & 0.31 & 0.74 & 0.31 & 0.06 & 0.31 & 0.12  \\
 \textsc{Comb} & \textbf{0.62} & \textbf{0.92} & \textbf{0.60} & \textbf{0.51} & 0.57 & \textbf{0.49} \\
 \end{tabular}
 \end{center}
}
\caption{Comparison between our best system (\textsc{Comb}), \newcite{jin-etal-2020-unsupervised}'s best system per-language (\textsc{PCS}) and their skyline (\textsc{CoNLL17}) on the inflection-from-lemma task.}
\label{tab:from_lemma}
\end{table}

\subsection{Results}

Table \ref{tab:main_res} summarizes the inflection accuracies for all models. Although the results vary widely across languages, \textsc{Comb} consistently achieves best or near-best results. While \textsc{Comb} outperforms the other two variants in Finnish, Hungarian and Turkish, its performance is roughly on par with \textsc{Orth} for the five Indo-European (IE) languages in our selection, pointing to the marginal role semantics played in those languages. Both \textsc{Comb} and \textsc{Orth} outperform \textsc{CA}, our stronger baseline.%

Comparing our best system to the \textsc{Sup} skyline, it seems that the room for improvement is bigger for languages with bigger paradigms (paradigm sizes are indicated in the table, both in number of forms and number of functions). Impressively, the results for English fall short only by $5$ percentage points from the fully-supervised upper-bound.

In terms of tagging accuracy, outputted data sets are quite invariably precise across most languages and models (details in the supplementary material). %
The success of the inflection model seem to be correlated with the amount of words tagged by the first module, rather than on it's precision. The Pearson correlation between the inflection accuracy and the log of the averaged tagged amount per paradigm cell is $0.87$.

To exemplify the added value of our minimally-supervised scenario we provide in Table \ref{tab:from_lemma} a comparison with the completely unsupervised model of \newcite{jin-etal-2020-unsupervised}. We compare their best model to our best model (\textsc{Comb}) applied on inflection from lemma. We provided our model with the same $100$ lemmas in their test set, and tested our model's capability to complete the respective inflection tables. For most languages, our model's inflection accuracy surpasses even the skyline named by \newcite{jin-etal-2020-unsupervised}, an edits-based minimally-supervised algorithm that uses $10$ inflection tables, while we are using only $5$. %

\subsection{Analysis}

\begin{table*}[t]
 \begin{center}
 \resizebox{\textwidth}{!}{
 \begin{tabular}{l|c|c|c|c|c|c|c|c|c}
 \textbf{Inflection} & \multirow{ 2}{*}{Average} & \multirow{ 2}{*}{\textsc{Eng}} & \multirow{ 2}{*}{\textsc{Rus}} & \multirow{ 2}{*}{\textsc{Lav}} & \multirow{ 2}{*}{\textsc{Fra}} & \multirow{ 2}{*}{\textsc{Fin}} & \multirow{ 2}{*}{\textsc{Spa}} & \multirow{ 2}{*}{\textsc{Tur}} & \multirow{ 2}{*}{\textsc{Hun}} \\
 \textbf{+hallucination}& & & & & & & & & \\
 \hline
 \hline
 \textsc{Overfit}  & 0.10 (+9) & 0.08 (+8) & 0.07 (+7) & 0.21 (+20) & 0.04 (+3) & 0.04 (+4) & 0.10 (+10) & 0.19 (+18) & 0.10 (+10) \\ 
 \textsc{CA}  & 0.34 (+16) & 0.46 (+8) & 0.40 (+10) & 0.41 (+26) & 0.22 (+16) & 0.24 (+18) & 0.21 (+19) & 0.38 (+16) & 0.40 (+17) \\
 \hline
 \textsc{Comb}  & 0.64 (+9) & 0.87 (-3) & 0.71 (-2) & 0.57 (+9) & 0.38 (+6) & 0.54 (+6) & 0.73 (+20) & 0.62 (+13) & 0.70 (+22)
 \end{tabular}
 }
 \end{center}
\caption{Results of the baselines and of our best model \textsc{Comb} when $5000$ hallucinated examples are added. In parenthesis are the differences comparing to the relevant result in Table \ref{tab:main_res}.}
\label{tab:hall_res}
\end{table*}

The results on our selection of languages, suggest a clear division between Indo-European (IE) languages and non-IE ones. In the former, adding semantic knowledge yields minute improvements at best, while in the latter, \textsc{Comb} clearly outperforms over \textsc{Orth}. We conjuncture that this is because the IE languages in our selection exhibit a fairly simple morpho-phonological system, with relatively few classes and almost no phonological stem-suffix interaction. In contrast, all non-IE languages selected exhibit vowel harmony that multiply the edits related to a single morphological relation, in addition to consonant gradation in the case of Finnish. 

This difficulty is magnified with a large amount of classes, as in the Finnish verbal paradigm that includes $27$ classes\footnote{according to the Research Institute for the Languages of Finland.} of which about a dozen seem to include more than a few lexemes according to the statistics over Wiktionary entries.\footnote{\url{https://en.wiktionary.org/wiki/Appendix:Finnish_conjugation}}

Another imbalance in the results is the inverse correlation between the size of the paradigm and the performance of the inflector trained over the  outputted data. We speculate that the effect arises from the fact that languages with bigger paradigms include inflections for functions that are in {\em exceedingly rare} use, either because they are considered archaic or literary, or because they are used for functions that are far less common. It means that the \textit{data-driven} vocabulary is likely to include less forms with those features, or miss them completely. It also means that these forms will have noisier vector embeddings.

\begin{figure}[t]
\begin{center}
\includegraphics[scale=0.5]{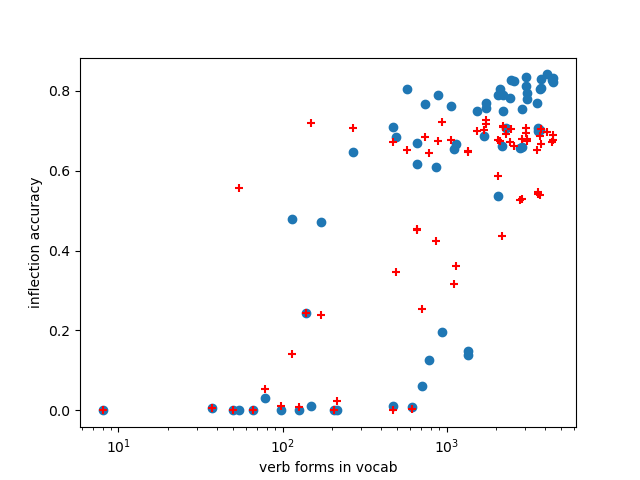}
\end{center}
\caption{Spanish inflection accuracy per morphological category as a function of the category's abundance in the vocabulary. Plotted results for \textsc{Orth} (\textcolor{red}{+}) and \textsc{Comb} (\textcolor{blue}{$\bullet$}).}
\label{fig:spa-break-down}
\end{figure}

\begin{figure}[t]
\begin{center}
\includegraphics[scale=0.22]{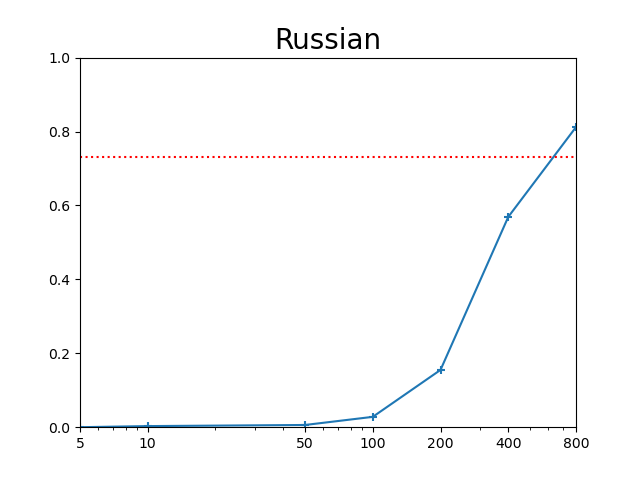}
\includegraphics[scale=0.22]{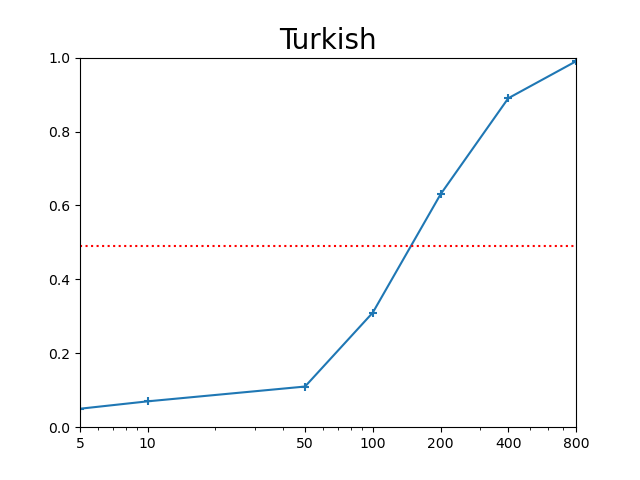}
\end{center}
\caption{Learning curves of the supervised model in $4$ of the languages: Russian and Turkish, with our best performance marked by the dashed line. Plots for all languages can be found in the supplementary material}
\label{fig:learning-curves}
\end{figure}

To probe this conjecture empirically, we plotted the Spanish inflection  by morphological category  against the number of forms found in FastText's vocabulary (Fig. \ref{fig:spa-break-down}). The figure includes results for both best systems in terms of inflection accuracy, namely \textsc{Orth} and \textsc{Comb}.
It shows that on common forms the inflection accuracy is on par with the performance on small-paradigm languages, while the rarer forms are mostly the ones driving the total accuracy down.\footnote{This analysis was possible only for Spanish as all other highly-inflectional languages have partial and skewed data on UniMorph, so automatically counting forms is impossible.}

In Fig. \ref{fig:learning-curves} we display the amount of annotation labor saved by using our model for a subset of the languages. We compare  the performance of \textsc{Comb} to models trained on increasing amounts of inflection tables, and we show that our model harnessed $5$ annotated inflection tables to output performance equivalent to over $400$ lexemes needed in a supervised scenario for most languages. That's a reduction of  two orders of magnitude in the labor needed for annotation. This may lead to more sophisticated annotation procedures that could capitalize on the bootstrapping approach proposed in this paper, to save time and efforts in creating morphological resources needed for many languages.

\paragraph{Data Hallucination} The methods introduced here to combat the annotation bottleneck are somewhat orthogonal to the method of data hallucination introduced in \citet{anastasopoulos-neubig-2019-pushing}, as we aim for identifying new \textit{real} examples from a vocabulary, rather than hallucinate nonce ones. Nonetheless we added $5000$ hallucinated examples to both baselines and to our best system (\textsc{Comb}) to assess the combined power of both methods (Table \ref{tab:hall_res}). Unsurprisingly, we found that hallucinated data helped improving the results of the baselines, since the hallucinated data is the sole source of new examples. However, this source of example is also quite noisy, and we see that for \textsc{Comb} adding hallucinated examples marginally harms the results for English and Russian, those languages with best \textsc{Comb} performance on our bootstrapped data. Furthermore, note that even with the hallucinated examples, the baselines still underperform compared to \textsc{Comb} models \textit{without} hallucination.

\begin{table*}[t]
 \begin{center}
 \resizebox{\textwidth}{!}{
 \begin{tabular}{l|c|c|c|c|c|c|c|c|c}
 \textbf{Inflection} & {Average} & {\textsc{Eng}} & {\textsc{Rus}} & {\textsc{Lav}} & {\textsc{Fra}} & {\textsc{Fin}} & {\textsc{Spa}} & {\textsc{Tur}} & {\textsc{Hun}} \\
 \hline
 \hline
 \textsc{Orth}  & 0.25 (-15) & 0.34 (-52) & 0.52 (-23) & 0.49 (+0) & 0.07 (-26) & 0.13 (+0) & 0.10 (-40) & 0.25 (-7) & 0.12 (+11) \\ 
 \textsc{Sem}  & 0.05 (-4) & 0.00 (-1) & 0.20 (-13) & 0.01 (+0) & 0.00 (-7) & 0.00 (+0) & 0.05 (-19) & 0.13 (+10) & 0.00 (+0) \\
 \textsc{Comb}  & 0.49 (-6) & 0.87 (-3) & 0.59 (-16) & 0.48 (+0) & 0.15 (-17) & 0.48 (+0) & 0.51 (-2) & 0.49 (+0) & 0.35 (-13)
 \end{tabular}
 }
 \end{center}
\caption{Results of our models when only frequency is considered in seed selection. In parenthesis are the differences comparing to the relevant result in Table \ref{tab:main_res}.}
\label{tab:seed_res}
\end{table*}

\paragraph{Error Analysis}
To  understand how it might be possible to improve the method further, we sampled $100$ incorrect examples from the  data set created using the \textsc{Comb} system for Spanish. We found that $64\%$ of the mistakes were of words tagged with only $1$ incorrect feature in the bundle. 
This suggests that a fine-grained algorithmic approach, that will tackle relations between individual features rather than complete sets, might do better in this regard. 

We examined the morpho-phonological patterns that appear in the outputs of both \textsc{Comb} and \textsc{Orth} for Finnish to better assess the reason for the gap in performance between them. We found that the data provided by \textsc{Comb} contains more examples for alternation unattested in the seed data comparing to \textsc{Orth}. For example, the data from \textsc{Comb} contained $19$ examples for the nt$\sim$nn alternation and $2$ examples for rt$\sim$rr, while the data from \textsc{Orth} contained no examples for both. For comparison, both datasets contained over $100$ examples for the attested t$\sim$tt alternation. We conclude that while both methods find examples for attested morpho-phonological processes, only \textsc{Comb} can close the gap on unattested processes and provide a more diverse dataset for better generalization.

\paragraph{Seed Selection}
As any bootstrapping method, our algorithms results' may be vulnerable to the selection of the minimal supervision set. To that end we purposefully aimed for frequent and diverse selection (Sec.~\ref{sec:exp_setup}). To examine the importance of the selection strategy we altered it to disregard class membership, and we automatically selected lexemes to maximize only frequency.
The results of this experiment are in Table~\ref{tab:seed_res}.

The results show that the selection procedure 
is indeed important as all different algorithms suffered a loss in performance. It is also evident that diversity is particularly crucial for the orthographic algorithm that is based on the different edit scripts and has less evidence when seed examples cover less classes.
\section{Related Work}
In recent years, neural sequence-to-sequence models have taken the  lead in all forms of morphological tagging and  morphological inflection tasks. 

Morphological tagging is an {\em analysis} task where the input is a complete sentence, i.e., a sequence of word forms, and the model aims to assign each  word form in-context a  morphological signature that consists of its lemma, part-of-speech (POS) tag, and a set of inflectional features \cite{hajic-2000-morphological, mueller-etal-2013-efficient, bohnet-etal-2018-morphosyntactic}. 

Morphological inflection works in the opposite direction, and may be viewed as a {\em generation} task. Here, forms of a lexeme are generated from one another given sets of inflectional features of both the input and output. In many implementations the input form is the lemma, in which case the inflectional features of the input are not given \cite{faruqui-etal-2016-morphological, aharoni-goldberg-2017-morphological}, and the lemma can be either spelled-out, or inputted as an index in a dictionary \cite{malouf-2017-abstractive}.\footnote{When inflection is done from a form other than the lemma, it is sometimes referred to as "reinflection".}

While most pioneering models for \textit{supervised} morphological inflection used statistical models based on finite-state-machines \cite{kaplan1994regular, eisner-2002-parameter}, nowadays neural models for morphological inflections are a lot more pervasive \cite{cotterell-etal-2016-sigmorphon, cotterell-etal-2017-conll, cotterell-etal-2018-conll} (and they go as back as \citealp{rumelhart-1986-learning}).

In the case of \textit{unsupervised} learning of morphology, a key task is to induce complete paradigms from unlabled texts. 
Early works on unsupervised morphology induction focused on morpheme segmentation for {\em concatenative} morphology \cite{goldsmith-2001-unsupervised, creutz-lagus-2002-unsupervised, narasimhan-etal-2015-unsupervised, bergmanis-goldwater-2017-segmentation}.
Notwithstanding, early unsupervised works that are {\em not} limited to concatenative morphology do exist \cite{yarowsky-wicentowski-2000-minimally}.

More recent studies on unsupervised morphology include works on knowledge transfer within a genealogical linguistic family well- to low-resourced languages \cite{kann-etal-2017-one, kann-etal-2020-learn, mccarthy-etal-2019-sigmorphon}, as well as works aimed at modifying 
the approaches for the supervised problem, to allow better
tackling of low resourced scenarios. These include \newcite{bergmanis-etal-2017-training} that focused on data augmentation, and \newcite{anastasopoulos-neubig-2019-pushing} that modified the model itself with a separate features encoder and introduced the now commonly-used hallucination method for data augmentation. The state of the art model for classic low-resourced scenarios, with a diverse but small dataset, is the transition-based model of \newcite{makarov-clematide-2018-neural}.
In addition, some works deal with other low-resourced scenarios and assume no inflection tables at all, and are focused on paradigm detection/completion in addition to inflection \cite{dreyer-eisner-2011-discovering, elsner-etal-2019-modeling, jin-etal-2020-unsupervised}. In contract, our scenario provides the knowledge on the paradigmatic structure with a small and undiverse supervision set.

Another work that made use of semantics is that of \newcite{soricut-och-2015-unsupervised}, who employed analogies between embedded words to \textit{filter} candidates affixation rules. We use embeddings to \textit{discover} examples in a more general morphological scenario.

\section{Conclusions}
\label{sec:conc}
In this work we propose a realistic minimally-supervised scenario for morphological inflection, which includes only a handful  of labeled inflection tables as well as a large bulk of unlabeled text. We showed that semantic and orthographic regularities allow bootstrapping the minimal supervision set to a large (noisy) labeled data set, by searching for word pairs in the vocabulary analogous to observed form pairs from the supervision. We demonstrate that training a neural morphological inflector over the bootstrapped dataset leads to some non-trivial successes, especially on paradigms of smaller size and on commonly-used inflections.
This contribution is orthogonal and can be applied in tandem with hallucination approaches. When applied separately, our method outperforms both hallucination and current state of the art models for low-resourced settings.
In the future we aim to improve performance over larger paradigms and rarer forms in order to make our method a viable substitute for the labor-intensive manual annotation for new languages.

\section*{Acknowledgements}
We thank Jonathan Berant for the helpful advice and discussion all throughout. We also thank the audience of the BIU-NLP seminar, the 18th SIGMORPHON meeting, and the 1st UniMorph Meeting, for comments and discussion. This research is funded by an ERC-StG Grant 677352 by the European Research Council, and an ISF grant 1739/26 by the Israeli Science Foundation, for which we are grateful.

\bibliography{anthology,custom}
\bibliographystyle{acl_natbib}

\appendix

\end{document}